\documentclass[letterpaper]{article} 
\usepackage{bm}
\usepackage{aaai25}  
\usepackage{times}  
\usepackage{helvet}  
\usepackage{courier}  
\usepackage[hyphens]{url}  
\usepackage{graphicx} 
\usepackage{natbib}  
\usepackage{caption} 
\usepackage{algorithm}
\usepackage{algorithmic}

\usepackage{amsmath}
\usepackage{multirow}
\usepackage{amssymb}
\usepackage{array}

\usepackage{newfloat}
\usepackage{listings}
\DeclareCaptionStyle{ruled}{labelfont=normalfont,labelsep=colon,strut=off} 
\floatstyle{ruled}
\newfloat{listing}{tb}{lst}{}
\floatname{listing}{Listing}
\title{UCF-Crime-DVS: A Novel Event-Based Dataset\\ for Video Anomaly Detection with Spiking Neural Networks}
\author{
    Yuanbin Qian\textsuperscript{\rm 1}\equalcontrib,
    Shuhan Ye\textsuperscript{\rm 1}\equalcontrib,
    Chong Wang\textsuperscript{\rm 1,\rm 2}\thanks{Corresponding author: Chong Wang.},
    Xiaojie Cai\textsuperscript{\rm 1},
    Jiangbo Qian\textsuperscript{\rm 1,\rm 2},
    Jiafei Wu\textsuperscript{\rm 3}
}
\affiliations{
    \textsuperscript{\rm 1}Faculty of Electrical Engineering and Computer Science, Ningbo University, China\\
    \textsuperscript{\rm 2}Merchants' Guild Economics and Cultural Intelligent Computing Laboratory, Ningbo University, China\\
    \textsuperscript{\rm 3}Department of Electrical and Electronic Engineering, The University of Hong Kong\\

    \{2311100301, 216002718, wangchong, 2211100083, qianjiangbo\}@nbu.edu.cn, jcjiafeiwu@gmail.com

}

\begin{document}

\maketitle

\begin{abstract}
Video anomaly detection plays a significant role in intelligent surveillance systems. To enhance model's anomaly recognition ability, previous works have typically involved RGB, optical flow, and text features. Recently, dynamic vision sensors (DVS) have emerged as a promising technology, which capture visual information as discrete events with a very high dynamic range and temporal resolution. It reduces data redundancy and enhances the capture capacity of moving objects compared to conventional camera. To introduce this rich dynamic information into the surveillance field, we created the first DVS video anomaly detection benchmark, namely UCF-Crime-DVS. To fully utilize this new data modality, a multi-scale spiking fusion network (MSF) is designed based on spiking neural networks (SNNs). This work explores the potential application of dynamic information from event data in video anomaly detection. Our experiments demonstrate the effectiveness of our framework on UCF-Crime-DVS and its superior performance compared to other models, establishing a new baseline for SNN-based weakly supervised video anomaly detection.

\end{abstract}

\begin{links}
    \link{Dataset and Code}{ https://github.com/YBQian-Roy/UCF-Crime-DVS}
\end{links}

\section{Introduction}

Video anomaly detection (VAD) is a crucial research direction in the fields of computer vision and machine learning, which plays a significant role in intelligent video surveillance system \cite{zhou2023dual}. For VAD tasks, content-rich datasets are effective in evaluating the strengths and weaknesses of algorithms and models. Benchmark datasets help define the scope of problems that can be solved. Some of the common publicly available benchmark datasets for VAD include UCSD-Peds \cite{li2013anomaly}, Avenue \cite{lu2013abnormal}, Street Scene \cite{ramachandra2020street}, Shanghai Tech \cite{luo2017revisit}, TAD \cite{lv2021localizing}, and UCF-Crime \cite{sultani2018real}, which cover various monitoring scenarios and anomalous events. Generally, these datasets are first processed through a feature extractor to obtain RGB features or optical flow features. RGB features capture the appearance information of the video, while optical flow features focus on the motion information.

\begin{figure}[t]
\centering
\includegraphics[width=1\columnwidth]{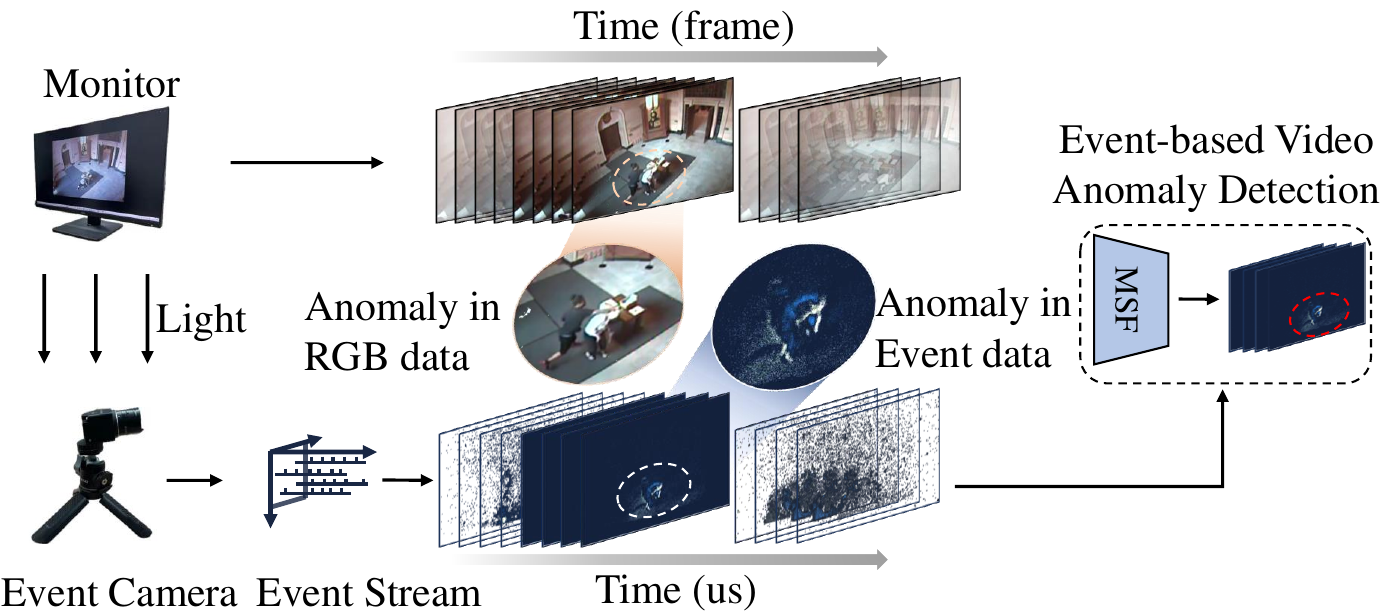} 
\caption{The overview of our contributions.}
\label{fig1}
\end{figure}

Recently, dynamic vision sensors (DVS) \cite{lichtsteiner2008128,brandli2014240}, also known as event cameras, have garnered significant attention due to their high dynamic range, high temporal resolution, and low latency. DVS is a bionic visual sensor inspired by human retinal peripheral neurons. It uses a difference-based sampling model to generate event data only when pixel brightness changes above a threshold. Unlike traditional images, event streams encode visual information as discrete events, dramatically reducing data redundancy and preserving temporal characteristics. This efficient information processing has enabled event cameras to capture moving objects in the frame better than conventional cameras and reduce system-level power consumption by up to 100 times \cite{delbruck2010activity,posch2010qvga}. However, despite their advantages, event cameras have not yet been applied to the field of VAD. Therefore, we introduced the first DVS dataset in this field using an event camera, called UCF-Crime-DVS, to explore the potential of it in VAD. 

\begin{table*}[!ht]
\centering

\begin{tabular}{lccccc}
\hline
\rule{0pt}{2ex} Dataset  & Sensors  & Class & Resolution &Sec Per Example &Object\\
\hline
N-Caltech101\cite{orchard2015converting}   & ATIS  & 101& 240 $\times$ 180 & 0.3s  &images\\
N-MNIST\cite{orchard2015converting}   & ATIS  & 10& 28 $\times$ 28 & 0.3s &images\\
CIFAR10-DVS\cite{li2017cifar10}   & DAVIS128  & 10& 128 $\times$ 128 & - &images\\
N-ImageNet\cite{kim2021n}   & Samsung-Gen3 & 1000 & 346 $\times$ 260 & - &images\\
ES-ImageNet\cite{lin2021imagenet}   & - & 1000 & 224 $\times$ 224 & - &images\\
DVS-Gesture\cite{amir2017low}   & DAVIS128 & 11 & 128 $\times$ 128 & 6s &action\\
N-CARS\cite{sironi2018hats}   & ATIS  & 2 & 128 $\times$ 128 & 0.1s&cars\\
ASL-DVS\cite{bi2019graph} & DAVIS240 & 24 & 346 $\times$ 260 & 0.1s&hand\\
ASLAN-DVS\cite{bi2020graph}   & DAVIS240c  & 432& 240 $\times$ 180 & - &action\\
HMDB-DVS\cite{bi2020graph}   & DAVIS240c  & 51 & 240 $\times$ 180 & 19s &action\\
UCF101-DVS\cite{bi2020graph}   & DAVIS240c  & 101 & 240 $\times$ 180 & 25s &action\\
PAF\cite{miao2019neuromorphic}   & DAVIS346 & 10 & 346 $\times$ 260 & 5s&action\\
DailyAction\cite{liu2021event}   & DAVIS346 & 12 & 346 $\times$ 260 & 5s&action\\
HARDVS\cite{wang2024hardvs}   & DAVIS346 & 300 & 346 $\times$ 260 & 5-10s&action\\
Bullying10K\cite{dong2024bullying10k} & DAVIS346 & 10 & 346 $\times$ 260
& 2-20s&action\\  \hline
\rule{0pt}{2ex} \textbf{UCF-Crime-DVS (Ours)}  &  IMX636 & 14 & \textbf{1280 $\times$ 720} &\textbf{avg 242s} &\textbf{anomaly}\\
\hline
\end{tabular}
\caption{Overview of various DVS datasets.}
\label{datasets}
\end{table*}

However, Artificial Neural Networks (ANNs) do not handle event streams well due to the discrete nature of event data. Unlike ANNs, Spiking Neural Networks (SNNs) receive event format data as input and use discrete, binary spike signals, leading to a natural advantage in handling event streams \cite{chen:23}. Therefore, to better utilize event data in the field of VAD, this paper introduces a fully SNN-based VAD framework called MSF. Given the unique dynamics and temporal complexity of event data, effectively processing this complexity is important. TIM \cite{shen2024tim} enhances the spiking self-attention (SSA) mechanism's ability to handle these challenges. Consequently, our MSF incorporates TIM to improve the model's temporal processing of event data.

To the best of our knowledge, this work pioneers the exploration of applying event data to VAD. The overview of our work is illustrated in Figure \ref{fig1}. First, we constructed an event-based dataset for VAD. With this dataset, we then present the MSF framework, a fully spiking neural network architecture designed to better detect anomalous events from event streams. Overall, our contributions can be summarized as follows:
\begin{itemize}
    \item We present the first large DVS dataset for VAD, in order to apply the rich dynamic information and high temporal resolution of DVS in VAD. 
    \item We propose a multi-scale SNN-based framework for DVS-based VAD. The Temporal Interaction Module (TIM) is innovatively incorporated in the convolution-based SNNs framework to enhance the integration of spiking features, demonstrating its effectiveness on other time-series tasks.  
\end{itemize}

\begin{figure*}[ht]
\centering
\includegraphics[width=0.8\textwidth]{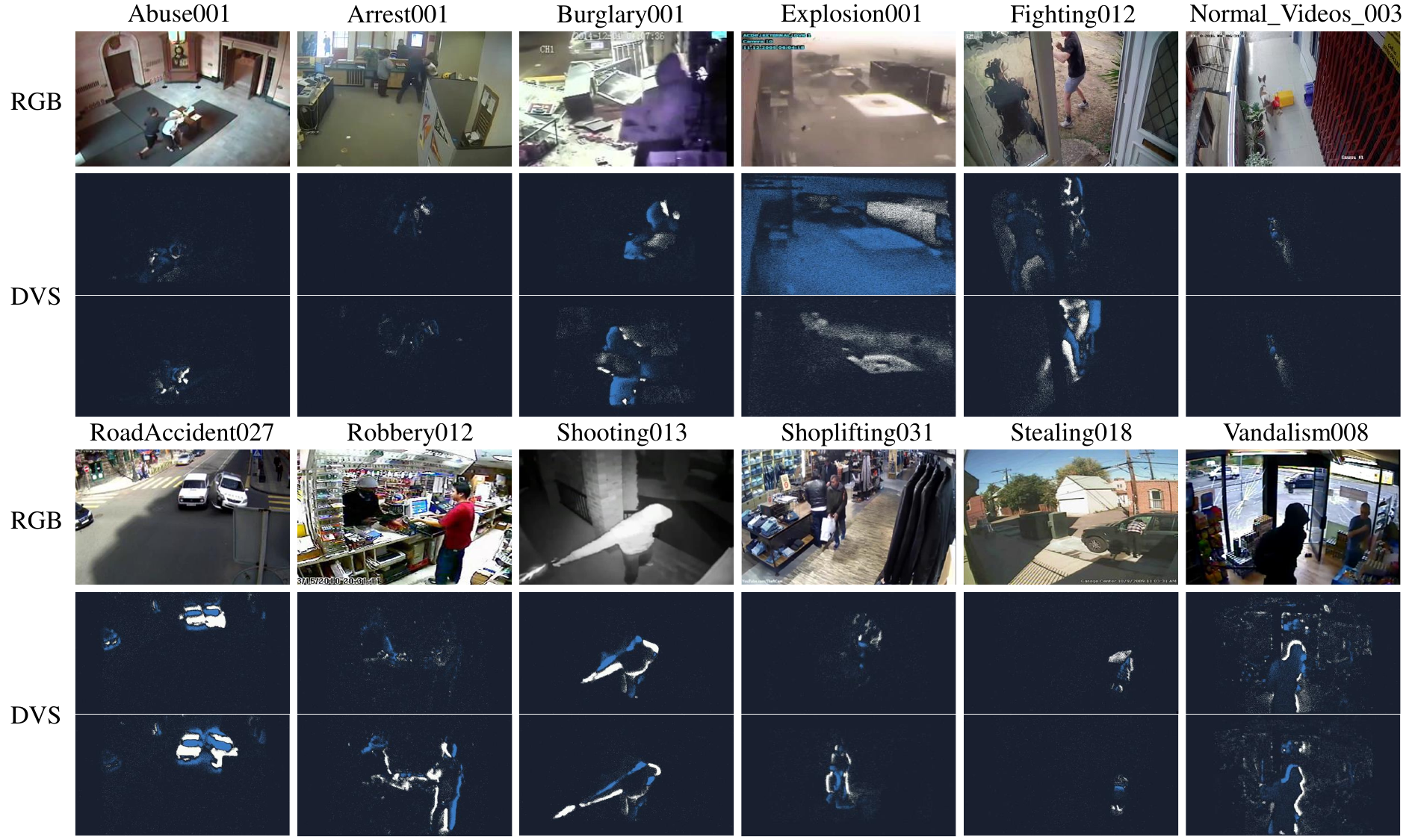} 
\caption{Presentation of our dataset and comparison between the DVS and RGB version of UCF-Crime.}
\label{fig2}
\end{figure*}

\section{Related Works}

\subsection{Event Camera Applications}

Event cameras have been widely used in computer vision applications. For example, TEF \cite{han2023high} reconstructs image signals by converting the high temporal resolution of the event signals into precise radiance values. SAN \cite{zhang2023generalizing} allows flexible input spatial scaling and uses self-supervised fine-tuning to enhance generalization performance for removing motion blur from images. STNet \cite{zhang2022spiking} dynamically extracts and fuses information from temporal and spatial domains for single-target tracking. ExACT \cite{zhou2024exact} introduces a novel approach to event-based action recognition by employing a cross-modal conceptualization. Although event cameras have been applied in many areas of computer vision, they have not yet been utilized in VAD. Therefore, our work explores this possibility.

\subsection{Weakly Supervised Video Anomaly Detection}

Our work is a weakly supervised video anomaly detection (WSVAD) task. The mainstream approach to it is multi-instance learning (MIL), proposed by \cite{sultani2018real}. Specifically, MIL treats each video as a "bag" and divides each video into equal-length, non-overlapping segments called instances. All instances in normal videos are called positive bags, while those that contain at least one abnormal instance are called negative bags, representing abnormal videos. In MIL, learning is performed by decreasing the predicted anomaly score for each instance in the positive bag and increasing the score only for the instance with the largest anomaly score in the negative bag. Overall, WSVAD can be summarized in three stages: 1) each video is segmented into multiple clips, and features are extracted by a pre-trained encoder; 2) anomaly scores are generated using a multilayer perceptron (MLP); 3) the model is optimized using the MIL framework.

\subsection{Spiking Neurons}

Since event cameras record the visual input as asynchronous discrete events, they are inherently suitable to cooperate with SNNs. Spiking neurons in forward propagation can be summarized in three steps: charge, fire, and reset \cite{fang2021deep}. In this paper, we choose the leaky integrate-and-fire (LIF) neuron model \cite{gerstner2014neuronal}, which is widely adopted in SNNs due to its simplicity and ability to capture key aspects of neuronal dynamics. The dynamic model of LIF can be written in the following form:
\begin{equation}\label{eq1}
    \mathbf{u}^{t+1,l} = \tau \mathbf{u}^{t,l} + \mathbf{W}^{l}\mathbf{o}^{t,l-1},
\end{equation}
\begin{equation}\label{eq2}
    \mathbf{o}^{t,l} = {\Theta}(\mathbf{u}^{t,l}-V_{th}),
\end{equation}
\begin{equation}\label{eq3}
    \mathbf{u}^{t+1,l} = \tau \mathbf{u}^{t,l} \cdot(1-\mathbf{o}^{t,l}) + \mathbf{W}^{l}\mathbf{o}^{t+1,l-1},
\end{equation}
where $\tau$ is leaky factor and $\bm{u}^{t,l}$ denotes membrane potential of the neurons in layer $l$ at time step $t$. 
$\mathbf{W}^{l}$ and $\mathbf{o}^{l}$ represent the weight parameters and the fired spikes, respectively. $\Theta$ denotes Heaviside step function. When $\mathbf{u}^{t,l}\geq V_{th}$ equals to one, otherwise equals to zero. The membrane potential accumulates with the input until a given threshold $V_{th}$ is exceeded, then the neuron delivers a spike and the membrane potential $\mathbf{u}^{t,l}$ is reset to zero.

\section{UCF-Crime-DVS Dataset}

For VAD, datasets are as fundamental as models. In our paper, we construct the first event-based VAD dataset, named UCF-Crime-DVS. Our dataset contains 1900 event streams across 13 anomaly classes, aligned with the original UCF-Crime dataset \cite{sultani2018real}. It includes 1610 training sets with video-level labels and 290 test sets with frame-level labels, maintaining an equal number of normal and abnormal videos in both. Table \ref{datasets} compares the parameters with other DVS dataset, highlighting that our dataset has a high resolution of 1280$\times$720 and an average duration of 242 seconds per video, totaling 128 hours. This far exceeds the specifications of other DVS datasets. Next, we will demonstrate the characteristics of the dataset and provide a detailed description of the dataset construction process.

\subsection{Characteristics of Event Data}

Unlike pixel points in RGB video, which has three channels $(red,green,blue)$, event data consists of only two channels $(OFF,ON)$. Specifically, each event can be represented by $\begin{aligned}e = (x,y,p,t)\end{aligned}$, where $(x,y)$ represents the position, $p \in \{0,1\}$ indicates the polarity, and $t$ represents the timestamp (microsecond, $\mu $s). Events where the brightness increases above the threshold are called $ON$ events, while those where the brightness decreases are called $OFF$ events. As mentioned above, the event camera uses a difference-based sampling model and a threshold mechanism to generate events. This mechanism allows event cameras to capture faster moving object and more dynamic information than RGB cameras while ignoring most static information. As shown in Figure \ref{fig2}, the dynamic subjects in our dataset are clearly presented, whereas the static background is barely visible. Additionally, small events at the frame edges, such as in Shoplifting031 and Stealing018, can be captured by the event camera.

\begin{figure}[t]
\centering
\includegraphics[width=1\columnwidth]{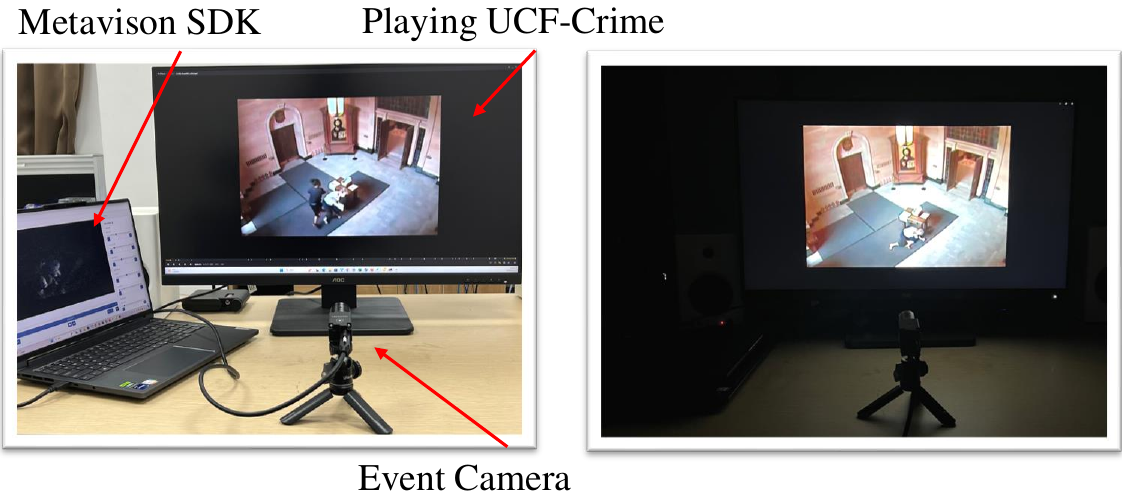} 
\caption{The final shooting environment setup.}
\label{envir}
\end{figure}

\subsection{Dataset Construction}

 \paragraph{Pre-Production Stage.}First of all, we prepared an event camera with a resolution of $1280\times 720$ and IMX636 sensors provided by Prophesee, and a 32" 4K monitor to play the original UCF-Crime dataset. The dataset was captured in a light-free environment, where the only light perceived by the event camera came from the monitor playing the videos.

 \paragraph{Dataset Pre-Processing Stage.}We combined the videos in the original dataset by class into single long videos for playback and recorded the number of frames in each video.

\begin{figure*}[t]
\centering
\includegraphics[width=1\textwidth]{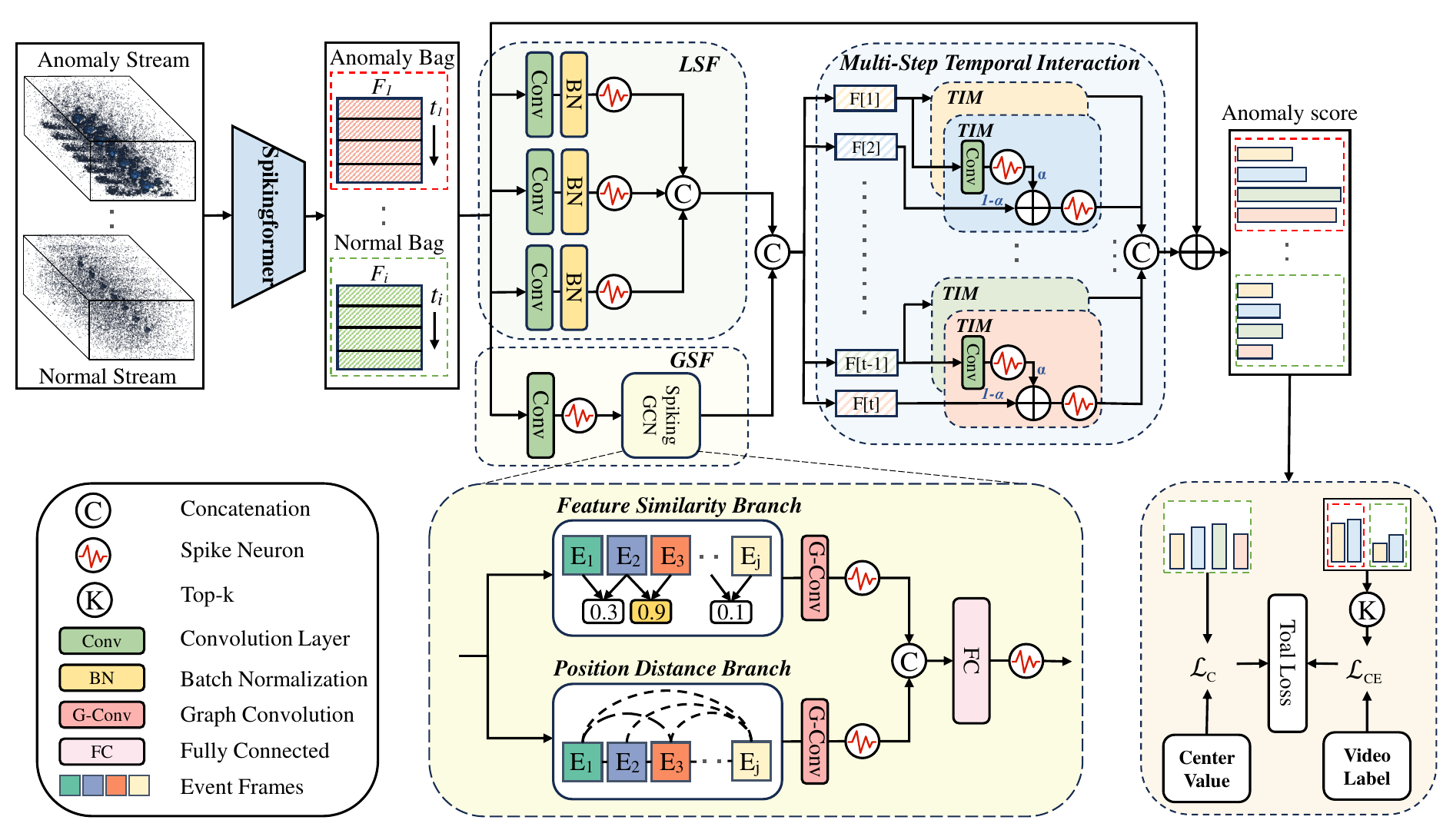} 
\caption{The framework of our proposed MSF. The LSF and GSF represents the local and global spiking feature extractor module, respectively. $\mathcal{L}_{\mathrm{CE}}$  denotes cross-entropy loss, and  $\mathcal{L}_{\mathrm{C}}$  denotes center loss.}
\label{fig3}
\end{figure*}

 \paragraph{Dataset Shooting Stage.} Metavision SDK is used to control the event camera. We adjusted the aperture and focus distance to capture sharp images. To reduce background noise, we fine-tuned the bias settings while following the event rate and the display to assess the noise impact. The final shooting setup is presented in Figure \ref{envir}.

 \paragraph{Dataset Post-Processing Stage.}

When the event dataset is recorded, we slice the long event stream by the length of each video segment, ensuring alignment with the original dataset. Since the discrete event data cannot be easily processed by the downstream networks, it need to be converted into a more usable format. The mainstream method integrates the event data into event frames based on the number of event frames or duration for downstream tasks. Similarly, we merged each event stream into event frames at designated time intervals. 

All events $e$ in every 533,328 $\mu $s (corresponding to 16 video frames) are integrated into an event frame $E_j$ which represents $j$-th event frame. Define $e_{\Delta t} = (x,y,p)$ as the event in $\Delta t$, where $\Delta t = t_{j_r}-t_{j_l}$. The process of integrating event can be expressed as:
\begin{equation}
    E_j(x,y,p)=\sum_{t = t_{j_l}}^{t_{j_r-1}}\mathbf{1}(e_{\Delta t}=(x_t,y_t,p_t)),
\end{equation}
here, $E_j(x,y,p)$ denotes the pixel value at position $(x,y,p)$ with which is integrated from the event data within the specified time interval $[t_{j_l},t_{j_r})$ and $\mathbf{1}$ is an indicator function that equals 1 only when $e_{\Delta t}=(x_t,y_t,p_t)$.

\section{Methods}
To effectively process the binary event streams dataset, we propose a multi-scale spiking fusion (MSF) network for WSVAD. Benefiting from the high temporal resolution and rich dynamic details of event data, the proposed multi-scale spiking fusion module can efficiently exploit temporal features. The complete structure of MSF is shown in Figure \ref{fig3}.

\subsection{Problem Statement }
Let $\mathbf{X} = \{\mathbf{x}_i\}_{i=1}^n$ represent the training set containing $n$ event stream videos from the proposed UCF-Crime-DVS dataset, and $\mathbf{T} = \{t_i\}_{i=1}^n$ denote the temporal duration, where $t_i$ is the event frame number of the $i$-th event stream. Additionally, we use $\mathbf{Y} = \{y_i\}_{i=1}^n$, where $y_i = \{0,1\}$, to represent the video anomaly label set. In the testing stage, the anomaly score vector for $i$-th video is defined as $\mathbf{s}_i =\{s^{j}\}_{j=1}^{t}$, where $s^{j} = \{0,1\}$, and $s^{j}$ is anomaly score of the $j$-th event clip.

\subsection{Feature Extraction}
Most VAD tasks start with feature extraction. We use Hardvs dataset \cite{wang2024hardvs}, a large event-based action recognition dataset, to pre-train a Spikingformer \cite{zhou2023spikingformer}, which serves as our feature extractor. The features of UCF-Crime-DVS are then extracted using the pre-trained Spikingformer. After that, we obtain the event stream feature $\mathbf{F}$ with dimensions $t \times D$ from the training video $\mathbf{x}$, where $D$ is the dimension of clip features. According to the multi-instance learning principle, the feature $\mathbf{F}$ is fed into our MSF. 

\subsection{Multi-Scale Spiking Fusion}
When dealing with event data, particularly for VAD, it is crucial to efficiently extract and retain temporal features while discovering temporal dependencies. Our proposed multi-scale spiking fusion module (MSF) captures both multi-resolution local spiking dependencies (light green block in Figure \ref{fig3}) within individual clip, and global spiking dependencies (light yellow block in Figure \ref{fig3}) between event clips. Finally, these dependencies are seamlessly integrated based on the unique characteristics of the spiking feature (light blue block in Figure \ref{fig3}).

\paragraph{Local Spiking Feature.} MSF uses pyramidal dilated convolution $\{\mathrm{P_1},\mathrm{P_2},\mathrm{P_3}\}$ over the temporal domain to learn multi-scale representations of event clips. It learns the multi-scale spiking features from the feature $\mathbf{F} = \{\mathbf{f}_d\}_{d=1}^D$. Given the feature $\mathbf{f}_d\in\mathbb{R}^{t}$, the one-dimensional dilated convolution operation is performed using the kernel $\mathbf{W}_{p,d}\in\mathbb{R}^{\omega}$ with $p\in\{1,...,D/4\}$, $d\in\{1,...,D\}$, and $\omega$ indicating the filter size, which is defined as:
\begin{equation}
    \mathbf{f}_p=\sum_{d=1}^D\mathbf{W}_{p,d}*\mathbf{f}_d,
\end{equation}
where $*$ denotes the dilated convolution operator, and $\mathbf{f}_p\in\mathbb{R}^{t}$ represents the output feature after applying dilated convolution in the time dimension. The features $\mathbf{F}_p \in \mathbb{R}^{{t}\times{D/4}}$that have been concatenated by $\mathbf{f}_p$ are then passed through spike neurons to obtain the spiking features:
\begin{equation}
    \mathbf{F}^\mathrm{P}=Lif(\mathbf{F}_p),
\end{equation}
where $Lif$ is the leaky integrate-and-fire spike neuron. 

\paragraph{Global Spiking Feature.} Despite of the local temporal dependencies, global ones are also important. We introduce a lightweight SpikingGCN to further capture the temporal dependencies across different event clips, which is shown in the yellowish green block in Figure \ref{fig3}. Our global temporal extraction module first downscales the features from $\mathbf{F}\in\mathbb{R}^{t\times D}$ to ${\mathbf{F}^{c}}\in\mathbb{R}^{t\times D/4}$ with ${\mathbf{F}^{c} = Conv_{1\times1} (\mathbf{F})}$. SpikingGCN then models global temporal dependencies of spiking feature in terms of feature similarity and relative distance. 

Feature similarity branch generates the adjacency matrix $\mathbf{M}_{sim}$ for SpikingGCN using event frame-wise cosine similarity method, which is denoted as follows,
\begin{equation}
\mathbf{M}^{sim}=\frac{\mathbf{F}^{c}\mathbf{F}^{{c}^\top}}{\left\|\mathbf{F}^{c}\right\|_2\cdot\left\|\mathbf{F}^{c}\right\|_2}.
\end{equation}

We employ the position distance branch to capture long-distance relationships between objects or scenes by measuring their positional differences across event frames, as illustrated below:
\begin{equation}\label{cc1}
    \mathbf{M}^{dis}(i,j)=\frac{-|i-j|}\sigma, 
\end{equation}
it means the proximity between event frames $i$ and $j$ depends solely  on their relative positions in time, independent of other factors. The hyperparameter $\sigma $ is used to adjust the degree of influence.

Overall, the modified SpikingGCN can be summarized as follows:
\begin{equation}
    {\mathbf{F}^{\mathrm{G}}}=Lif\left(\left[Soft(\mathbf{M}^{sim});Soft(\mathbf{M}^{dis})\right]\mathbf{F}^{c}\mathbf{W}\right),
\end{equation}
where $\mathbf{W}$ is the unique learnable weight used to transform the input feature space into another feature space. $Soft$ indicates the Softmax normalization, which is used to ensure the sum of each row of $\mathbf{M}^{sim}$ and $\mathbf{M}^{dis}$ equals to one.

\begin{table*}[!ht]
    \centering

\begin{tabular}{lcccc}
\hline
\rule{0pt}{2ex} 
Methods & Architecture &
Supervision & 
           AUC(\%)         & FAR(\%)         \\ \hline
Sultani et al. \cite{sultani2018real}                                            & ANNs         & Weakly-supervised                                     & 55.56             & 8.69             \\
3C-Net \cite{narayan20193c}                                            & ANNs              & Weakly-supervised                                  & 59.22              & 9.50              \\
AR-Net \cite{wan2020weakly}                                       & ANNs               & Weakly-supervised                                 & 60.71           & 8.51           \\
Wu et al. \cite{wu2020not}                                           & ANNs           & Weakly-supervised                                     & 58.58              & 34.35              \\

RTFM \cite{tian2021weakly}                                         & ANNs             & Weakly-supervised                                   & 52.67              & 13.19              \\
TSA \cite{joo2023clip}                                   & ANNs                    & Weakly-supervised                            & 51.86              & 22.36              \\  
\hline
SEW-ResNet \cite{fang2021deep}                                  & SNNs            & Weakly-supervised                                    & 53.99              & 11.79              \\  
PLIF \cite{fang2021incorporating}                                  & SNNs            & Weakly-supervised                                    & 54.74              & 9.17              \\  
baseline\cite{zhou2023spikingformer}                                   & SNNs            & Weakly-supervised                                    & 62.78              & 11.52              \\  
\hline 
\rule{0pt}{2ex} 
\textbf{MSF(Ours)}                        & SNNs  & Weakly-supervised                                     & \textbf{65.01}           & \textbf{3.27}            \\ \hline
\end{tabular}
    \caption{AUC and FAR of the proposed method against other methods on UCF-Crime-DVS. These methods are adapted to our achitecture and re-trained on the UCF-Crime-DVS.}
\label{table2}
\end{table*}

\paragraph{Multi-Scale Spiking Interaction.}We use residual concatenation to prevent features from being over-smoothed and concatenate global spiking features with local spiking features, which can be describe as,
\begin{equation}
    \bar{\mathbf{F}}=[\mathbf{F}^{(l)}]_{l\in L}\in\mathbb{R}^{t\times D},
\end{equation}
where $L=\{\mathrm{P}_1,\mathrm{P}_2,\mathrm{P}_3,\mathrm{G}\}$. $\mathbf{F}^\mathrm{P}$ and $\mathbf{F}^\mathrm{G}$ refer to the learned local and global temporal features respectively.

As previously mentioned, event data possesses unique dynamics and temporal intricacies, whereas the membrane potentials of spiking neurons exhibit a cumulative nature. Therefore, the extracted temporal information initially manifests as membrane potentials rather than spikes. Consequently, traditional ANN-based temporal learning methods, such as MTN \cite{tian2021weakly}, fail to effectively integrate the multi-scale features of event clips, leading to substantial under-utilization of information from different time steps. To exploit the hidden information across various time steps, we employ the Temporal Interaction Module (TIM) \cite{shen2024tim} to fuse historical spike information with current spike information. The hyperparameter $\alpha $ is used as a weight parameter, allowing the model to balance the combination of historical states and current inputs during computation. This can be mathematically expressed by the following equation:
\begin{equation}\label{cc2}
    \mathbf{F}^{\mathrm{TIM}}=\alpha Conv(\mathbf{F}^{\mathrm{TIM}}[t-1])+(1-\alpha)\bar{\mathbf{F}}[t].
\end{equation}

TIM demonstrates a dual mechanism for temporal information processing: immediate feature extraction and historical state integration. This approach not only extracts key features from the current input but also effectively utilizes the implicit state information from previous time steps. This design achieves an organic combination of short-term and long-term dependencies, enabling the model to capture the complex dynamics in event data.

\paragraph{Anomaly Scorer.}After MSF, a fully connected (FC) layer and a sigmoid function are employed as an anomaly scorer to generate the anomaly score vector $\mathbf{s_i}$:
\begin{equation}
    \mathbf{s_i}=Sigmoid(FC(\mathbf{F}^{\mathrm{TIM}})).
\end{equation}

\begin{table}[!t]
    \centering
    
\begin{tabular}{ccc|cc}
\hline
\rule{0pt}{2ex} 
LSF & GSF & TIM & AUC(\%) & FAR(\%)\\ \hline
-&-    & -    & 62.78   & 11.52 \\
$\checkmark$    &  -   &  -   & 62.44   & 5.06\\
   -  & $\checkmark$   &   -  & 60.32   & 6.80\\
  -   &  -   & $\checkmark$   & 50.06   & \textbf{0.68}\\
$\checkmark$    & $\checkmark$   &  -   & 55.69  & 7.27 \\
$\checkmark$    &   -  & $\checkmark$   & 64.39  & 2.80 \\
  -   & $\checkmark$   & $\checkmark$   & 64.07  & 5.13 \\ \hline
\rule{0pt}{2ex} 
$\checkmark$    & $\checkmark$   & $\checkmark$   & \textbf{65.01}  & 3.27\\ \hline

\end{tabular}
\caption{Ablation study for different module.}
\label{table3}
\end{table}

\subsection{Loss Function}

The classic dynamic multiple-instance learning (DMIL) loss and center loss are implemented for our proposed MSF.

\paragraph{DMIL Loss.}
The DMIL loss is to enlarge the inter-class distance of instances, which can be represented as follows:
\begin{equation}
    \begin{aligned}\mathcal{L}_{\mathrm{DMIL}}&=\frac{1}{k_{i}}\sum_{s_{i}^{j}\in \mathbf{S}_{i}}[-y_{i}log(s_{i}^{j})\\&+(1-y_{i})log(1-s_{i}^{j})]\end{aligned},
\end{equation}
where $\mathbf{s}_{i}^{j}$ is descending sorted anomaly score vector of the $i$-th video, $\mathbf{S}_{i}=\{s_{i}^{j}\mid j=1,2,...,k_{i}\}$ consists of top-$k_i$ elements in $s_i$ and $y_{i}=\{0,1\}$ is the video anomaly label.

\paragraph{Center Loss.} The center loss used for anomaly score regression collects the anomaly scores of normal event clips, reducing the intra-class distance. It can be represented as,
\begin{equation}
    \mathcal{L}_c=\begin{cases}\frac{1}{t_i}\sum\limits_{j=1}^{t_i}\left\|s_i^j-c_i\right\|_2^2,&\text{if~} y_i=0\\\\0,&\text{otherwise}\end{cases},
\end{equation}
\begin{equation}
    c_i=\frac1{t_i}\sum_{j=1}^{t_i}s_i^j,
\end{equation}
where $c_i$ is the center of anomaly score vector $s_i$.
Overall, the total loss function of our MSF model is given by:
\begin{equation}\label{cc3}
    \mathcal{L}=\mathcal{L}_{\mathrm{DMIL}}+\lambda \mathcal{L}_{c}.
\end{equation}

\begin{figure}[!t]
\centering
\includegraphics[width=1\columnwidth]{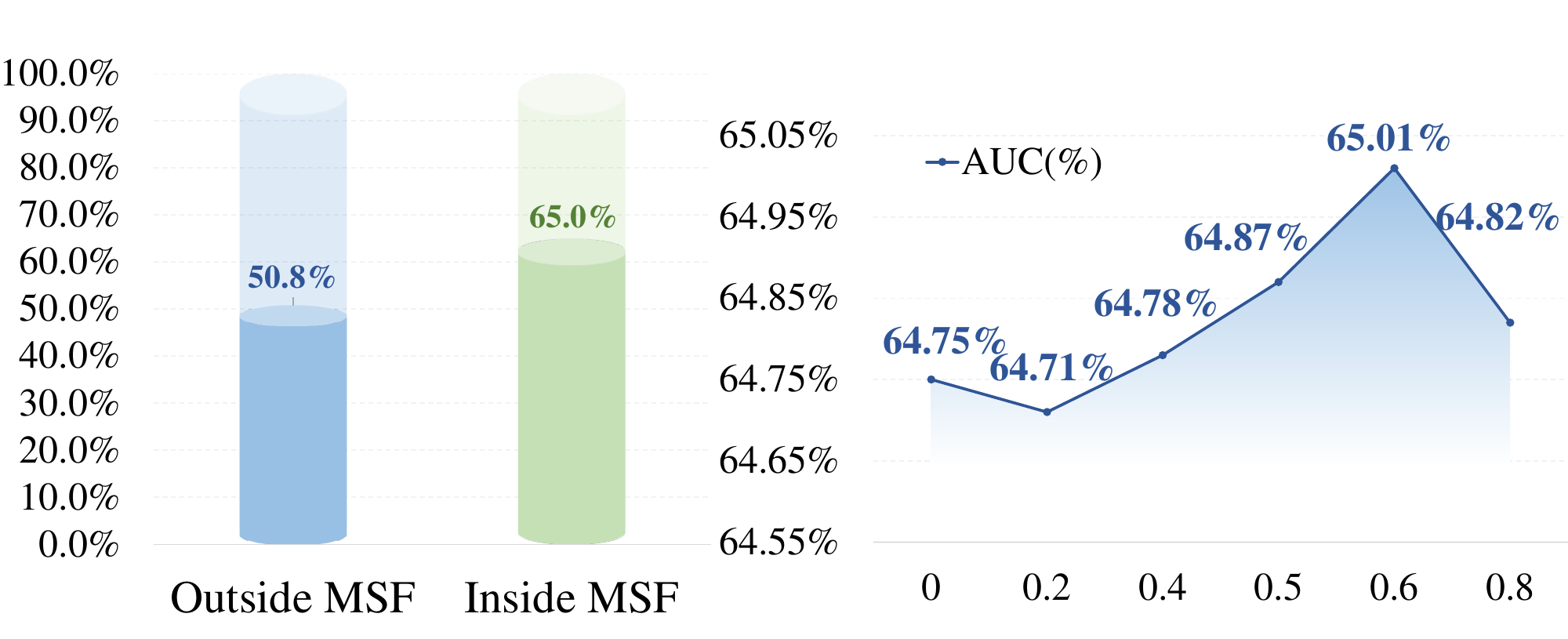} 
\caption{Ablation study for TIM. Left: Performance comparison with different position of TIM. Right: Performance comparison with different $\alpha$  values.}
\label{tim}
\end{figure}

\begin{table}[!t]
\centering
\setlength{\tabcolsep}{1mm}
\begin{tabular}{lccccccc}
\hline
\rule{0pt}{2ex} 
\textbf{$\tau$} & 0.2 & 0.25 & 0.4 & 0.5 & 0.625 & 0.8 \\
\hline
\rule{0pt}{2ex}
AUC(\%) & 63.99 & 63.96 & 64.37 & 64.78 & \textbf{65.01} & 64.00 \\
\hline
\end{tabular}
\caption{Performance comparison with different time constant $\tau$ on UCF-Crime-DVS.}
\label{table7}
\end{table}

\section{Experiments }

We validated our UCF-Crime-DVS dataset and MSF framework using a VAD task. Additionally, we tested the ability of each module with ablation experiments.

\subsection{Experiments Setup}
\paragraph{Training Dataset.} We use our UCF-Crime-DVS dataset to test and verify our proposed method. Our UCF-Crime-DVS dataset is aligned with the UCF-Crime dataset, covering 13 classes of anomalies in 1,610 training videos with video-level labels and 290 test videos with frame-level labels.

\begin{figure}[!t]
\centering
\includegraphics[width=1\columnwidth]{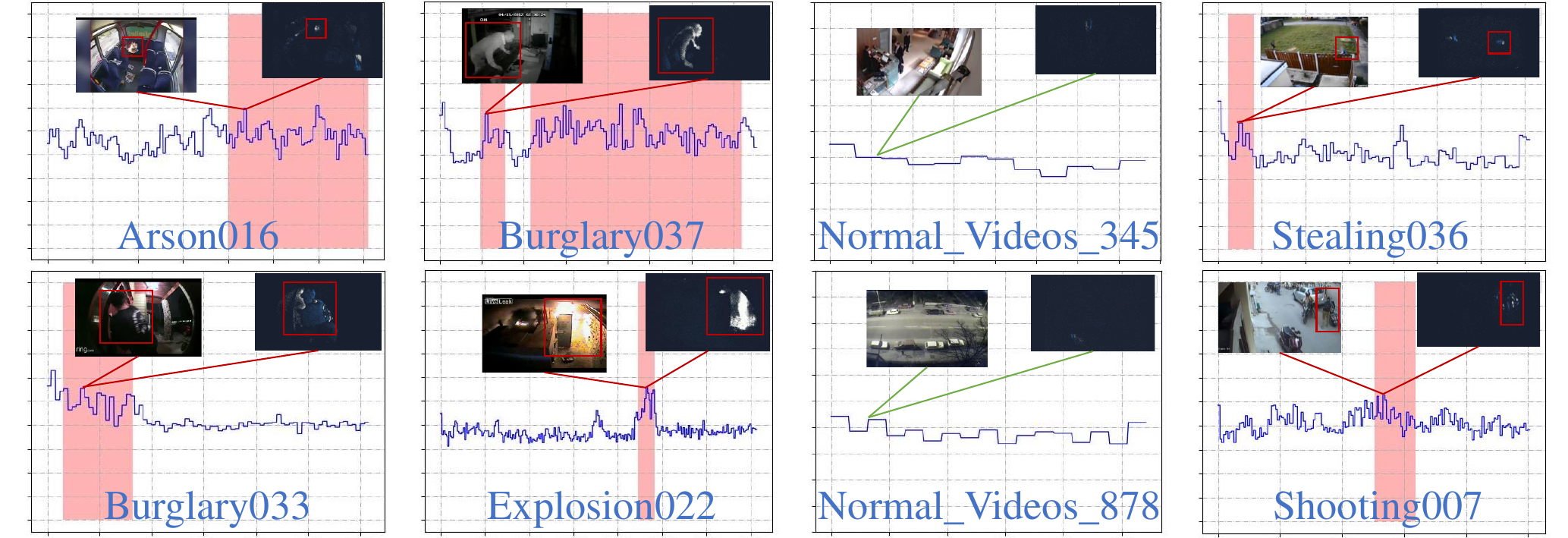} 
\caption{Anomaly scores of our methods on UCF-Crime-DVS. Pink areas indicate the manually labelled abnormal events, purple lines represent the anomaly score and red boxes point out abnormal events on the screen.}
\label{fig4}
\end{figure}

\paragraph{Training Details.}Following \cite{sultani2018real}, each event stream is divided into non-overlapped clips. Empirically, we set $k=4$ for our dataset. We use the Adam optimizer with a weight decay of 0.0005, and a learning rate of 0.0001. For $\sigma$ in Eq.\ref{cc1} and $\lambda$ in Eq.\ref{cc3}, we set them as 1 and 20, respectively. Each batch contains 60 samples, split equally between normal and abnormal video sequences, which are randomly selected from the training set. The models for conducting experiments are implemented based on Pytorch \cite{paszke2019pytorch}, SpikingJelly \cite{fang2023spikingjelly} and a server with single RTX4090 GPU. 

\paragraph{Evaluation Metrics.}We use two standardized performance metrics to evaluate the anomaly detection capability of the model: Area Under of Curve (AUC) of the frame-level Receiver Operating Characteristics (ROC) and False Alarm Rate (FAR) with a threshold 0.5. The combined assessment of these two metrics not only reflects the overall discriminative ability of the model, but also its reliability and stability in real-world application scenarios. 

\subsection{Performance Analysis} 

Table \ref{table2} presents a comparison of our method against other methods on UCF-Crime-DVS dataset. It can be seen that compared classical VAD frameworks do not perform well on this dataset. Some methods have a FAR of more than 20\%, suggesting they are unable to process event data effectively. Other SNN-based architectures with deeper network layers also fail to achieve a high AUC and low FAR at the same time, indicating that simply increasing the network complexity does not improve the VAD performance. Our MSF, on the other hand, achieves an AUC of 65.01\% for anomaly detection, along with a FAR of only 3.27\%, which has 3\% more AUC and 8\% lower FAR than our baseline. It establishes a new baseline for event-based WSVAD. 

\subsection{Ablation Study}

A series of ablation studies presented in Table \ref{table3} demonstrate that optimal performance is achieved when all three modules are combined. In contrast, the performance of the LSF-GSF combination, as well as each module individually, is suboptimal. This can be attributed to the fact that both LSF and GSF expand feature representations in the temporal domain, with the LSF lacking interconnections and the GSF smoothing features, which imposes a challenge for anomaly localization. However, integrating the TIM module with LSF and GSF significantly improves performance, highlighting the TIM module's critical role in effectively integrating information across time steps.

\paragraph{Ablations for TIM.}As shown in the left of Figure \ref{tim}, we conducted ablation experiments to examine the impact of TIM placement. The results reveal a more than 10\% accuracy difference between the two placements, indicating that the optimal placement of TIM is within the MSF module. Integrating TIM within MSF enables seamless fusion of temporal features across multiple time steps, ensuring accurate capture of temporal dependencies and improving anomaly detection performance. Additionally, as seen in the right of Figure \ref{tim}, any non-zero value for $\alpha$ yields better results than setting $\alpha$ to zero. MSF achieves its best performance when $\alpha$ is set to 0.6, indicating that the introduction of temporal interaction significantly enhances performance.

\paragraph{Ablations for Time Constant.}A smaller time constant $\tau$ results in more leakage of the membrane potential over time, potentially leading to a loss of temporal information. To optimize the model, we performed ablation experiments on the time constants $\tau$. Table \ref{table7} shows that the model's ability to detect anomalies initially increases with $\tau$, reaching a peak accuracy of 65.01\% when $\tau$ is set to 0.625. However, when $\tau$ exceeds 0.625, AUC begins to decline, dropping further at 0.8. This suggests that an excessively large $\tau$ reduces the model's temporal memory capacity. Therefore, it is indispensable to select an appropriate $\tau$ through experimentation.

\subsection{Visualization}

A set of visualization are presented in Figure \ref{fig4}. Noting that certain scene transitions and the opening or closing credits exhibit similar characteristics to the explosion events, e.g. flickering visuals and a surge in event occurrences. It makes detecting explosion events in our dataset very challenging. However, our model still successfully recognizes explosion events such as Explosion022, highlighting its robustness. Additionally, for subtle anomalous events like stealing, which are difficult to detect visually, our model is able to identify these weak anomalies to some extent, as illustrated by the case of Stealing036. Although the visualized anomaly scores do not consistently exceed the anomaly threshold in anomalous segments, this is because some anomalous events include relatively stationary segments that did not trigger DVS, resulting in partial loss of these events.

\section{Conclusion}
In this paper, we present the first event-based VAD dataset and introduce the MSF framework for SNN-based VAD. Extensive experiments demonstrate that our method outperforms others on UCF-Crime-DVS, highlighting its potential for real-world applications. While our method has not yet achieved such high accuracy of traditional approaches on RGB dataset, it offers a fresh perspective on VAD and lays the foundation for future research.

\section{Acknowledgments}
This work was supported by the Ningbo Municipal Natural Science Foundation of China (No. 2022J114), National Natural Science Foundation of China (No. 62271274), Ningbo S\&T Project (No.2024Z004) and Ningbo Major Research and Development Plan Project (No.2023Z225).

\bibliography{aaai25}

\begin{thebibliography}{40}
\providecommand{\natexlab}[1]{#1}

\bibitem[{Amir et~al.(2017)Amir, Taba, Berg, Melano, McKinstry, Di~Nolfo, Nayak, Andreopoulos, Garreau, Mendoza et~al.}]{amir2017low}
Amir, A.; Taba, B.; Berg, D.; Melano, T.; McKinstry, J.; Di~Nolfo, C.; Nayak, T.; Andreopoulos, A.; Garreau, G.; Mendoza, M.; et~al. 2017.
\newblock A low power, fully event-based gesture recognition system.
\newblock In \emph{Proceedings of the IEEE conference on computer vision and pattern recognition}, 7243--7252.

\bibitem[{Bi et~al.(2019)Bi, Chadha, Abbas, Bourtsoulatze, and Andreopoulos}]{bi2019graph}
Bi, Y.; Chadha, A.; Abbas, A.; Bourtsoulatze, E.; and Andreopoulos, Y. 2019.
\newblock Graph-based object classification for neuromorphic vision sensing.
\newblock In \emph{Proceedings of the IEEE/CVF international conference on computer vision}, 491--501.

\bibitem[{Bi et~al.(2020)Bi, Chadha, Abbas, Bourtsoulatze, and Andreopoulos}]{bi2020graph}
Bi, Y.; Chadha, A.; Abbas, A.; Bourtsoulatze, E.; and Andreopoulos, Y. 2020.
\newblock Graph-based spatio-temporal feature learning for neuromorphic vision sensing.
\newblock \emph{IEEE Transactions on Image Processing}, 29: 9084--9098.

\bibitem[{Brandli et~al.(2014)Brandli, Berner, Yang, Liu, and Delbruck}]{brandli2014240}
Brandli, C.; Berner, R.; Yang, M.; Liu, S.-C.; and Delbruck, T. 2014.
\newblock A 240$\times$ 180 130 db 3 $\mu$s latency global shutter spatiotemporal vision sensor.
\newblock \emph{IEEE Journal of Solid-State Circuits}, 49(10): 2333--2341.

\bibitem[{Chen et~al.(2023)Chen, Peng, Li, and Tian}]{chen:23}
Chen, G.; Peng, P.; Li, G.; and Tian, Y. 2023.
\newblock Training full spike neural networks via auxiliary accumulation pathway.
\newblock \emph{arXiv preprint arXiv:2301.11929}.

\bibitem[{Delbr{\"u}ck et~al.(2010)Delbr{\"u}ck, Linares-Barranco, Culurciello, and Posch}]{delbruck2010activity}
Delbr{\"u}ck, T.; Linares-Barranco, B.; Culurciello, E.; and Posch, C. 2010.
\newblock Activity-driven, event-based vision sensors.
\newblock In \emph{Proceedings of 2010 IEEE international symposium on circuits and systems}, 2426--2429. IEEE.

\bibitem[{Dong et~al.(2024)Dong, Li, Zhao, Shen, and Zeng}]{dong2024bullying10k}
Dong, Y.; Li, Y.; Zhao, D.; Shen, G.; and Zeng, Y. 2024.
\newblock Bullying10K: a large-scale neuromorphic dataset towards privacy-preserving bullying recognition.
\newblock \emph{Advances in Neural Information Processing Systems}, 36.

\bibitem[{Fang et~al.(2023)Fang, Chen, Ding, Yu, Masquelier, Chen, Huang, Zhou, Li, and Tian}]{fang2023spikingjelly}
Fang, W.; Chen, Y.; Ding, J.; Yu, Z.; Masquelier, T.; Chen, D.; Huang, L.; Zhou, H.; Li, G.; and Tian, Y. 2023.
\newblock Spikingjelly: An open-source machine learning infrastructure platform for spike-based intelligence.
\newblock \emph{Science Advances}, 9(40): eadi1480.

\bibitem[{Fang et~al.(2021{\natexlab{a}})Fang, Yu, Chen, Huang, Masquelier, and Tian}]{fang2021deep}
Fang, W.; Yu, Z.; Chen, Y.; Huang, T.; Masquelier, T.; and Tian, Y. 2021{\natexlab{a}}.
\newblock Deep residual learning in spiking neural networks.
\newblock \emph{Advances in Neural Information Processing Systems}, 34: 21056--21069.

\bibitem[{Fang et~al.(2021{\natexlab{b}})Fang, Yu, Chen, Masquelier, Huang, and Tian}]{fang2021incorporating}
Fang, W.; Yu, Z.; Chen, Y.; Masquelier, T.; Huang, T.; and Tian, Y. 2021{\natexlab{b}}.
\newblock Incorporating learnable membrane time constant to enhance learning of spiking neural networks.
\newblock In \emph{Proceedings of the IEEE/CVF international conference on computer vision}, 2661--2671.

\bibitem[{Gerstner et~al.(2014)Gerstner, Kistler, Naud, and Paninski}]{gerstner2014neuronal}
Gerstner, W.; Kistler, W.~M.; Naud, R.; and Paninski, L. 2014.
\newblock \emph{Neuronal dynamics: From single neurons to networks and models of cognition}.
\newblock Cambridge University Press.

\bibitem[{Han et~al.(2023)Han, Asano, Shi, Zheng, and Sato}]{han2023high}
Han, J.; Asano, Y.; Shi, B.; Zheng, Y.; and Sato, I. 2023.
\newblock High-fidelity event-radiance recovery via transient event frequency.
\newblock In \emph{Proceedings of the IEEE/CVF Conference on Computer Vision and Pattern Recognition}, 20616--20625.

\bibitem[{Joo et~al.(2023)Joo, Vo, Yamazaki, and Le}]{joo2023clip}
Joo, H.~K.; Vo, K.; Yamazaki, K.; and Le, N. 2023.
\newblock Clip-tsa: Clip-assisted temporal self-attention for weakly-supervised video anomaly detection.
\newblock In \emph{2023 IEEE International Conference on Image Processing (ICIP)}, 3230--3234. IEEE.

\bibitem[{Kim et~al.(2021)Kim, Bae, Park, Zhang, and Kim}]{kim2021n}
Kim, J.; Bae, J.; Park, G.; Zhang, D.; and Kim, Y.~M. 2021.
\newblock N-imagenet: Towards robust, fine-grained object recognition with event cameras.
\newblock In \emph{Proceedings of the IEEE/CVF international conference on computer vision}, 2146--2156.

\bibitem[{Li et~al.(2017)Li, Liu, Ji, Li, and Shi}]{li2017cifar10}
Li, H.; Liu, H.; Ji, X.; Li, G.; and Shi, L. 2017.
\newblock Cifar10-dvs: an event-stream dataset for object classification.
\newblock \emph{Frontiers in neuroscience}, 11: 309.

\bibitem[{Li, Mahadevan, and Vasconcelos(2013)}]{li2013anomaly}
Li, W.; Mahadevan, V.; and Vasconcelos, N. 2013.
\newblock Anomaly detection and localization in crowded scenes.
\newblock \emph{IEEE transactions on pattern analysis and machine intelligence}, 36(1): 18--32.

\bibitem[{Lichtsteiner, Posch, and Delbruck(2008)}]{lichtsteiner2008128}
Lichtsteiner, P.; Posch, C.; and Delbruck, T. 2008.
\newblock A 128$\times $128 120 dB 15$\mu $ s latency asynchronous temporal contrast vision sensor.
\newblock \emph{IEEE journal of solid-state circuits}, 43(2): 566--576.

\bibitem[{Lin et~al.(2021)Lin, Ding, Qiang, Deng, and Li}]{lin2021imagenet}
Lin, Y.; Ding, W.; Qiang, S.; Deng, L.; and Li, G. 2021.
\newblock Es-imagenet: A million event-stream classification dataset for spiking neural networks.
\newblock \emph{Frontiers in neuroscience}, 15: 726582.

\bibitem[{Liu et~al.(2021)Liu, Xing, Tang, Ma, and Pan}]{liu2021event}
Liu, Q.; Xing, D.; Tang, H.; Ma, D.; and Pan, G. 2021.
\newblock Event-based Action Recognition Using Motion Information and Spiking Neural Networks.
\newblock In \emph{IJCAI}, 1743--1749.

\bibitem[{Lu, Shi, and Jia(2013)}]{lu2013abnormal}
Lu, C.; Shi, J.; and Jia, J. 2013.
\newblock Abnormal event detection at 150 fps in matlab.
\newblock In \emph{Proceedings of the IEEE international conference on computer vision}, 2720--2727.

\bibitem[{Luo, Liu, and Gao(2017)}]{luo2017revisit}
Luo, W.; Liu, W.; and Gao, S. 2017.
\newblock A revisit of sparse coding based anomaly detection in stacked rnn framework.
\newblock In \emph{Proceedings of the IEEE international conference on computer vision}, 341--349.

\bibitem[{Lv et~al.(2021)Lv, Zhou, Cui, Xu, Li, and Yang}]{lv2021localizing}
Lv, H.; Zhou, C.; Cui, Z.; Xu, C.; Li, Y.; and Yang, J. 2021.
\newblock Localizing anomalies from weakly-labeled videos.
\newblock \emph{IEEE transactions on image processing}, 30: 4505--4515.

\bibitem[{Miao et~al.(2019)Miao, Chen, Ning, Zi, Ren, Bing, and Knoll}]{miao2019neuromorphic}
Miao, S.; Chen, G.; Ning, X.; Zi, Y.; Ren, K.; Bing, Z.; and Knoll, A. 2019.
\newblock Neuromorphic vision datasets for pedestrian detection, action recognition, and fall detection.
\newblock \emph{Frontiers in neurorobotics}, 13: 38.

\bibitem[{Narayan et~al.(2019)Narayan, Cholakkal, Khan, and Shao}]{narayan20193c}
Narayan, S.; Cholakkal, H.; Khan, F.~S.; and Shao, L. 2019.
\newblock 3c-net: Category count and center loss for weakly-supervised action localization.
\newblock In \emph{Proceedings of the IEEE/CVF international conference on computer vision}, 8679--8687.

\bibitem[{Orchard et~al.(2015)Orchard, Jayawant, Cohen, and Thakor}]{orchard2015converting}
Orchard, G.; Jayawant, A.; Cohen, G.~K.; and Thakor, N. 2015.
\newblock Converting static image datasets to spiking neuromorphic datasets using saccades.
\newblock \emph{Frontiers in neuroscience}, 9: 437.

\bibitem[{Paszke et~al.(2019)Paszke, Gross, Massa, Lerer, Bradbury, Chanan, Killeen, Lin, Gimelshein, Antiga et~al.}]{paszke2019pytorch}
Paszke, A.; Gross, S.; Massa, F.; Lerer, A.; Bradbury, J.; Chanan, G.; Killeen, T.; Lin, Z.; Gimelshein, N.; Antiga, L.; et~al. 2019.
\newblock Pytorch: An imperative style, high-performance deep learning library.
\newblock \emph{Advances in neural information processing systems}, 32.

\bibitem[{Posch, Matolin, and Wohlgenannt(2010)}]{posch2010qvga}
Posch, C.; Matolin, D.; and Wohlgenannt, R. 2010.
\newblock A QVGA 143 dB dynamic range frame-free PWM image sensor with lossless pixel-level video compression and time-domain CDS.
\newblock \emph{IEEE Journal of Solid-State Circuits}, 46(1): 259--275.

\bibitem[{Ramachandra and Jones(2020)}]{ramachandra2020street}
Ramachandra, B.; and Jones, M. 2020.
\newblock Street scene: A new dataset and evaluation protocol for video anomaly detection.
\newblock In \emph{Proceedings of the IEEE/CVF Winter Conference on Applications of Computer Vision}, 2569--2578.

\bibitem[{Shen et~al.(2024)Shen, Zhao, Shen, and Zeng}]{shen2024tim}
Shen, S.; Zhao, D.; Shen, G.; and Zeng, Y. 2024.
\newblock TIM: An Efficient Temporal Interaction Module for Spiking Transformer.
\newblock \emph{arXiv preprint arXiv:2401.11687}.

\bibitem[{Sironi et~al.(2018)Sironi, Brambilla, Bourdis, Lagorce, and Benosman}]{sironi2018hats}
Sironi, A.; Brambilla, M.; Bourdis, N.; Lagorce, X.; and Benosman, R. 2018.
\newblock HATS: Histograms of averaged time surfaces for robust event-based object classification.
\newblock In \emph{Proceedings of the IEEE conference on computer vision and pattern recognition}, 1731--1740.

\bibitem[{Sultani, Chen, and Shah(2018)}]{sultani2018real}
Sultani, W.; Chen, C.; and Shah, M. 2018.
\newblock Real-world anomaly detection in surveillance videos.
\newblock In \emph{Proceedings of the IEEE conference on computer vision and pattern recognition}, 6479--6488.

\bibitem[{Tian et~al.(2021)Tian, Pang, Chen, Singh, Verjans, and Carneiro}]{tian2021weakly}
Tian, Y.; Pang, G.; Chen, Y.; Singh, R.; Verjans, J.~W.; and Carneiro, G. 2021.
\newblock Weakly-supervised video anomaly detection with robust temporal feature magnitude learning.
\newblock In \emph{Proceedings of the IEEE/CVF international conference on computer vision}, 4975--4986.

\bibitem[{Wan et~al.(2020)Wan, Fang, Xia, and Mei}]{wan2020weakly}
Wan, B.; Fang, Y.; Xia, X.; and Mei, J. 2020.
\newblock Weakly supervised video anomaly detection via center-guided discriminative learning.
\newblock In \emph{2020 IEEE international conference on multimedia and expo (ICME)}, 1--6. IEEE.

\bibitem[{Wang et~al.(2024)Wang, Wu, Jiang, Bao, Zhu, Li, Wang, and Tian}]{wang2024hardvs}
Wang, X.; Wu, Z.; Jiang, B.; Bao, Z.; Zhu, L.; Li, G.; Wang, Y.; and Tian, Y. 2024.
\newblock Hardvs: Revisiting human activity recognition with dynamic vision sensors.
\newblock In \emph{Proceedings of the AAAI Conference on Artificial Intelligence}, volume~38, 5615--5623.

\bibitem[{Wu et~al.(2020)Wu, Liu, Shi, Sun, Shao, Wu, and Yang}]{wu2020not}
Wu, P.; Liu, J.; Shi, Y.; Sun, Y.; Shao, F.; Wu, Z.; and Yang, Z. 2020.
\newblock Not only look, but also listen: Learning multimodal violence detection under weak supervision.
\newblock In \emph{Computer Vision--ECCV 2020: 16th European Conference, Glasgow, UK, August 23--28, 2020, Proceedings, Part XXX 16}, 322--339. Springer.

\bibitem[{Zhang et~al.(2022)Zhang, Dong, Zhang, Ding, Heide, Yin, and Yang}]{zhang2022spiking}
Zhang, J.; Dong, B.; Zhang, H.; Ding, J.; Heide, F.; Yin, B.; and Yang, X. 2022.
\newblock Spiking transformers for event-based single object tracking.
\newblock In \emph{Proceedings of the IEEE/CVF conference on Computer Vision and Pattern Recognition}, 8801--8810.

\bibitem[{Zhang et~al.(2023)Zhang, Yu, Yang, Liu, and Xia}]{zhang2023generalizing}
Zhang, X.; Yu, L.; Yang, W.; Liu, J.; and Xia, G.-S. 2023.
\newblock Generalizing event-based motion deblurring in real-world scenarios.
\newblock In \emph{Proceedings of the IEEE/CVF International Conference on Computer Vision}, 10734--10744.

\bibitem[{Zhou et~al.(2023)Zhou, Yu, Zhou, Ma, Zhang, Zhou, and Tian}]{zhou2023spikingformer}
Zhou, C.; Yu, L.; Zhou, Z.; Ma, Z.; Zhang, H.; Zhou, H.; and Tian, Y. 2023.
\newblock Spikingformer: Spike-driven residual learning for transformer-based spiking neural network.
\newblock \emph{arXiv preprint arXiv:2304.11954}.

\bibitem[{Zhou, Yu, and Yang(2023)}]{zhou2023dual}
Zhou, H.; Yu, J.; and Yang, W. 2023.
\newblock Dual memory units with uncertainty regulation for weakly supervised video anomaly detection.
\newblock In \emph{Proceedings of the AAAI Conference on Artificial Intelligence}, volume~37, 3769--3777.

\bibitem[{Zhou et~al.(2024)Zhou, Zheng, Lyu, and Wang}]{zhou2024exact}
Zhou, J.; Zheng, X.; Lyu, Y.; and Wang, L. 2024.
\newblock ExACT: Language-guided Conceptual Reasoning and Uncertainty Estimation for Event-based Action Recognition and More.
\newblock In \emph{Proceedings of the IEEE/CVF Conference on Computer Vision and Pattern Recognition}, 18633--18643.

\end{thebibliography}

\end{document}